\documentclass[9pt,nonacm,sigplan]{acmart}

\usepackage{soul, xcolor}
\usepackage{comment}

\AtBeginDocument{%
  \providecommand\BibTeX{{%
    \normalfont B\kern-0.5em{\scshape i\kern-0.25em b}\kern-0.8em\TeX}}}

\setcopyright{acmcopyright}
\copyrightyear{2023}
\acmYear{2023}
\acmDOI{XXXXXXX.XXXXXXX}

\acmConference[ICONS '23]{International Conference of Neuromorphic Systems}{August 01--03,
  2023}{Santa Fe, NM}
\acmPrice{15.00}
\acmISBN{978-1-4503-XXXX-X/18/06}

\setcopyright{none}
\renewcommand\footnotetextcopyrightpermission[1]{}
\begin{document}

\title{Zespol: A Lightweight Environment for Training Swarming Agents}


\author{Shay Snyder, Kevin Zhu, Ricardo Vega, Cameron Nowzari, Maryam Parsa}
\affiliation{%
  \institution{George Mason University}
  \city{Fairfax}
  \state{VA}
  \country{USA}}
\email{{ssnyde9, kzhu4, rvega7, cnowzari, mparsa}@gmu.edu}






\renewcommand{\shortauthors}{Snyder, et al.}

\begin{abstract}
Agent-based modeling (ABM) and simulation have emerged as important tools for studying emergent behaviors, especially in the context of swarming algorithms for robotic systems. 
Despite significant research in this area, there is a lack of standardized simulation environments, which hinders the development and deployment of real-world robotic swarms. 
To address this issue, we present Zespol, a modular, Python-based simulation environment that enables the development and testing of multi-agent control algorithms. Zespol provides a flexible and extensible sandbox for initial research, with the potential for scaling to real-world applications. 
We provide a topological overview of the system and detailed descriptions of its plug-and-play elements. We demonstrate the fidelity of Zespol in simulated and real-word robotics by replicating existing works highlighting the simulation to real gap with the milling behavior.
We plan to leverage Zespol's plug-and-play feature for neuromorphic computing in swarming scenarios, which involves using the modules in Zespol to simulate the behavior of neurons and their connections as synapses. This will enable optimizing and studying the emergent behavior of swarm systems in complex environments. Our goal is to gain a better understanding of the interplay between environmental factors and neural-like computations in swarming systems.

\end{abstract}





\maketitle

\section{Introduction}
Despite having limited individual capabilities, species such as bees \cite{harshey1994bees} and ants~\cite{garnier2007biological}, combine their abilities collectively to achieve impressive feats, such as honeybee house hunting~\cite{garnier2007biological} and weaver ant nest construction~\cite{garnier2007biological}. Researchers have drawn inspiration from this natural phenomenon and developed weak agents capable of working together to solve unique challenges, including clustering \cite{tzanetos2020comprehensive, brown2018discovery}, classification \cite{tzanetos2020comprehensive}, self-triggered coordination \cite{NOWZARI20121077}, and asynchronous cloud access \cite{nowzariMultiagentCoordinationAsynchronous2016}. However, there are several issues hindering the advancement of emergent behaviors in low-powered, disposable robotic swarms. The \ul{first} issue is the lack of standardization in simulation, development, and evaluation, resulting in challenging verification and extension of results~\cite{Brown2016DiscoveryAE, https://doi.org/10.48550/arxiv.2301.09018}. The \ul{second} issue involves domain adaptation problems when trying to recreate simulated emergent behaviors physically, which leads to significant performance reductions~\cite{HH-TA-AR:20, GV-CV-GS-TN-AEE-TV:18, https://doi.org/10.48550/arxiv.2301.09018}. \ul{Finally}, the increasing complexity of robotic control algorithms~\cite{9199280} present a significant obstacle to simulating multiple physical models required for swarming robotics. Although robotics research has often focused on developing individual capabilities \cite{adate2022survey}, simulating swarming robotics requires the simulation of multiple physical models. However, current examples of distributed robotic simulation environments are often designed for larger-scale manufacturing and reinforcement learning problems \cite{DBLP:journals/corr/abs-1810-05762}, have complicated C and C++ interfaces \cite{DBLP:journals/corr/abs-1810-05762}, or are specialized for particular applications \cite{soria2020swarmlab}.

  We developed Zespol, a Python-based simulation environment, to overcome the challenges of engineering swarms of robots with emergent behaviors. Zespol aims to facilitate research into the application of nature-inspired computing algorithms for swarming robotics. It provides a lightweight means of standardization for engineering swarms of robotics to emerge at collective behaviors before transitioning over to real-world experiments or higher fidelity simulations. Additionally, this framework establishes a direct connection to other simulation environments, minimizing domain adaptation penalties when switching from simulation to physical robotic systems. Zespol has native support for distributed parallelization and provides a modular, extensible, and well-documented Python-based interface compatible with neuromorphic computing platforms.

\section{Background \& Motivation}
There have been a variety of previous works aimed at creating simulation environments for robotics and swarming applications. In \cite{bettini2022vmas}, the Vectorized Multi-Agent Simulator (VMAS) is introduced as a framework designed for efficient multi-agent reinforcement learning applications. Their physics engine and the underlying control logic is written in PyTorch. VMAS has shown incredible performance in allowing parallel environments to run on GPUs. While this focus on CUDA accelerated hardware is beneficial for GPU-compatible workloads, it struggles when the simulation environment needs to expand to multiple heterogeneous compute nodes. A core assumption of the VMAS platform is holonomicity. We believe emergent behaviors can be decoupled from the sensing and control problem. Zespol does not force this trade-off of simulation fidelity for speed, as this would compromise the connection between simulation and a non-holonomic real robot.


Swarm-Sim as a 2D \& 3D simulation core for swarm agents stands as a framework for the implementation and evaluation of swarming agents \cite{cheraghi2020swarm}.
The major limitations of this framework are the lack of direct support for cluster-level parallelization, and the discrete grid coordinate system, which precludes emergent behaviors that depend on agents having a continuous state, such as milling with ground robots.

Introduced in \cite{9340854}, SwarmLab is a MATLAB \cite{MATLAB:2010} based drone swarm simulator. The simulator is designed around a Drone class that supports quadcopter and fixed-wing aircraft dynamics. This makes their framework unsuitable for use with any non-drone agents without extensive modifications to the code. Interfacing MATLAB with modern learning and neuromorphic processing frameworks presents additional difficulties. There are also notable performance issues associated with this framework that make running large-scale simulations a computationally expensive task.


MASON \cite{luke2005mason} is an agent-based simulation library designed from the ground-up to support custom Java-based \cite{arnold2005java} simulations.
There are many similarities between MASON and Zespol such as the inherent separation between the environment and visualization systems along with the compartmentalized nature of individual simulations. Both MASON and Zespol allow agents to be given arbitrary dynamics. The major limitation of MASON is a consequence of using very advanced Java where the barrier to entry for new users can be high. This issue is only compounded when we consider the lack of a mature and low-barrier system for distributing these simulations among heterogeneous computing systems. Addressing these issues is one of the major goals of Zespol.


OpenAI Gym, introduced in 2016, was a pioneering platform in the field of single-agent reinforcement learning \cite{DBLP:journals/corr/BrockmanCPSSTZ16}.
Out of the box, they support a wide variety of applications for classic control problems such as Box2D \cite{catto2011box2d}, and Atari \cite{bellemare2013arcade}. Compared to Zespol, Gym has two major limitations in that it is primarily designed for reinforcement learning and the programmatic architecture around Gym is focused purely on single agent simulations which severely limits its applicability to multi-agent robotics~\cite{https://doi.org/10.48550/arxiv.2301.09018, majid2022swarm}.

NetLogo \cite{Netlogo} is another multi-agent simulation environments. It is primarily designed to be used in educational environments, as evidenced by its integrated IDE with a drag-and-drop GUI. This makes programming behaviors easy, but the NetLogo language is limited. It is possible to run Python and R code from within NetLogo, as well as invoke a NetLogo simulation from a Java environment, but the interfaces are clunky and limited; thus NetLogo is largely incompatible with current means of distributing computation and simulation environments among heterogeneous computing systems and modern learning frameworks. NetLogo's simulation speed is, at best, equal to that of MASON \cite{luke2005mason}, but struggles with anything higher than two-dimensional environments.

In \cite{https://doi.org/10.48550/arxiv.2301.09018}, they conducted an interactive simulation in the design loop where simulated experiments where tightly coupled with real-world experiments. This study was broken up into four distinct portions to minimize the simulation to reality gap:
1) Characterizing the salient capabilities of the real robot, 2) Building a minimally viable simulation environment that characterizes the measured capabilities of physical robots, 3) Developing and exploring potential emergent behaviors in simulation, and 4) Deploying real robots based on simulation-driven-hypothesis and evaluating the performance penalties associated with the domain shift. They used a binary controller \cite{FB-MG-RN:21} for the salient capabilities of real robots and created stable milling behaviors in NetLogo that also performed the same behavior on physical robots. Despite their ability to minimize the simulation to reality gap, we are interested in deploying low-power and scalable neuromorphic computing platforms and explore novel methods of arriving at emergent behaviors. Zespol is designed as a simulation framework compatible with existing neuromorphic frameworks \cite{8573122, 10.1145/3381755.3381758, Schuman_Kulkarni_Parsa_Mitchell_Date_Kay_2022} and hardware \cite{10.1145/3381755.3381764}. 

Some of the key differences between prior works and Zespol are summarized in Table \ref{tab:sim_table}. Zespol is the only simulator that is written in user-friendly and well documented Python code, provides native capability for distributed (dist) simulation environments, and allows for arbitrary agent states and dynamics.

\begin{table}[h]
\caption{Comparison of multi-agent simulators}
\begin{tabular}{ r l l l l l}
    \textbf{Simulator} & \textbf{Lang} & \textbf{Dist} & \textbf{State} & \textbf{Dynamics} \\
    \hline \hline
    VMAS~\cite{bettini2022vmas}        & Python  & No & Arbitrary  & Holonomic \\
    Swarm-Sim~\cite{cheraghi2020swarm} & Python  & No & Discrete   & Holonomic \\
    SwarmLab~\cite{9340854}            & MATLAB  & No & Continuous & Drone     \\
    MASON~\cite{luke2005mason}         & Java    & No & Arbitrary  & Arbitrary \\
    NetLogo~\cite{Netlogo}             & NetLogo & No & Arbitrary  & Arbitrary \\
    Gym~\cite{DBLP:journals/corr/BrockmanCPSSTZ16}       & Python  & No & Arbitrary  & Arbitrary \\
    \hline
    \textbf{Zespol}    & \textbf{Python}  & \textbf{Yes} & \textbf{Arbitrary}  & \textbf{Arbitrary} \\
    \end{tabular}
    \label{tab:sim_table}
\end{table}

\section{Programmatic Architecture}

Zespol's underlying architecture is designed with modularity in mind where each fundamental building block has a plug and play interface. This design philosophy allows users to develop their own blocks such as sensor modules, controllers, and physical dynamics. All simulations are designed to minimize inter-object dependencies to reduce the chance of segmentation faults and minimize communication latency by only passing critical information between blocks. Each interface is thoroughly documented with the provided examples showing how users can extend the framework to support their needs. More formally, each building block is represented by two data structures that form an \textbf{object-state} relationship. We have provided three fundamental object-state pairs, Agent-AgentState, Swarm-SwarmState, and World-WorldState. A more detailed description of these pairs is given in the following.



\begin{figure}[h]
    \includegraphics[width=0.4\textwidth]{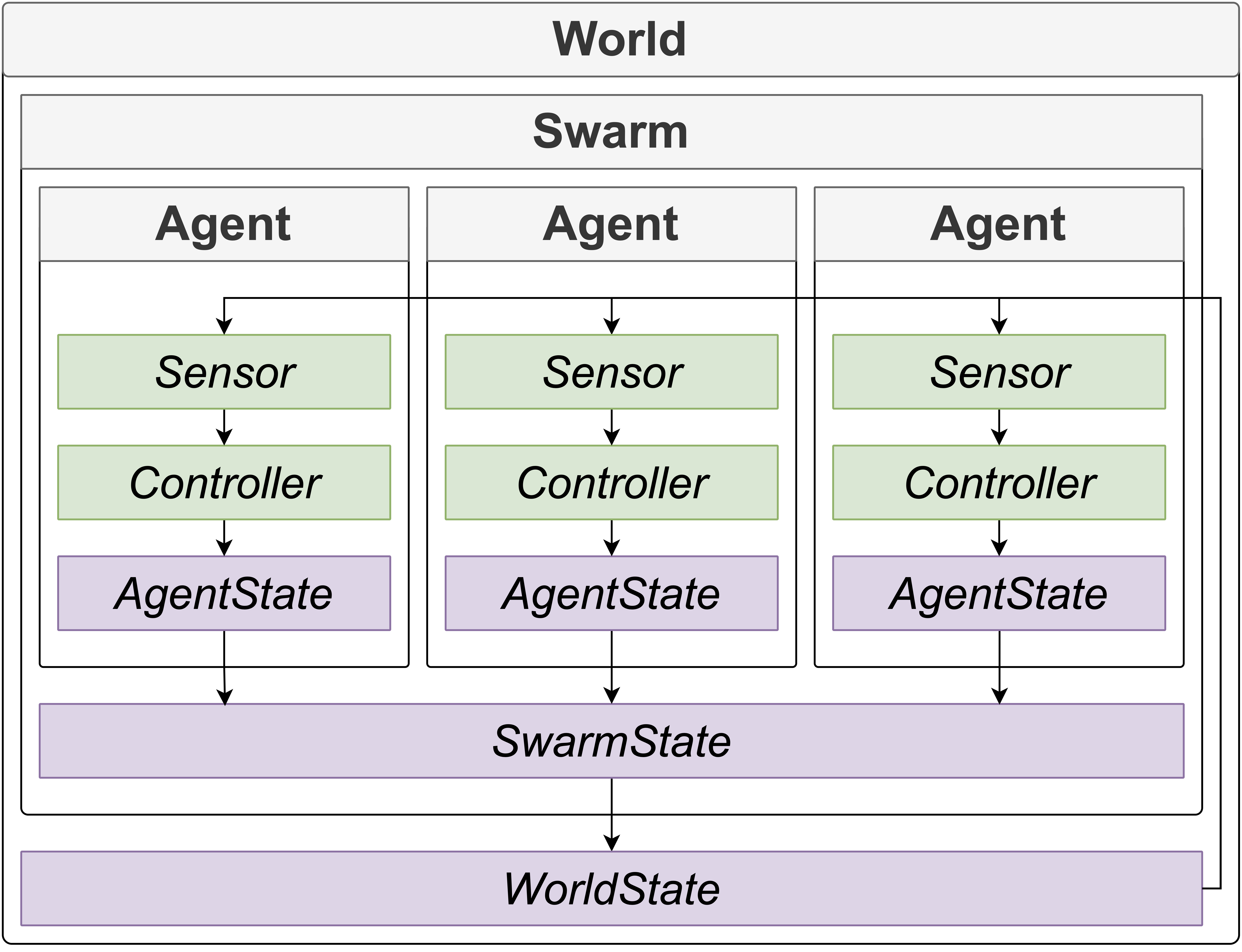}
    \caption{A flowchart presenting the critical programmatic flow within and between Zespol's main components.}
    \label{fig:program_arch}
\end{figure}

Zespol objects are data structures responsible for containing all elements required for the object to function. For example, a robot object would contain the robot's current location, all of its sensor objects, controller objects, and control the interaction between these elements at every simulation time step.

The \textit{\textbf{``Agent"}} object base class should be extended to support the specific requirements of a user's application. For example, The base class defines position and orientation vectors along three dimensions, a unique identifier, and the simulation time step fidelity. However, the \textit{tick} method must be updated based on user requirements to control the interactions between sensors, controllers, and physical dynamics.


The \textit{\textbf{``Swarm"}} class contains references to all agents within the swarm and controls the interactions between agents at every simulation time step. This is where the distributed nature of Zespol is highlighted because the memory and process spaces for all agents are separated, the processing of individual agent updates at every time step can be distributed across heterogeneous compute clusters with tools such as Dask \cite{dask}.


Bringing everything together, we have the \textbf{\textit{``World"}} class that contains every object and actionable element within the simulation environment. Therefore, this object maintains references to all swarms, visualization systems, and environmental objects such as world boundaries and obstacles. The last major responsibility of World objects is to manage the interactions between all swarms and environmental objects to manage the programmatic flow at every simulation time step.

For every agent, swarm, and world object there are associated states that contains a holistic view of the object with the central idea being the establishment of a shareable data structure that only contains fundamental information. This avoids repeatedly passing redundant information between objects. For example, AgentStates contain an Agent's location and orientation but shouldn't contain a copy of the Agent's sensor or controller.

\textbf{\textit{``AgentsStates"}} are defined by a snapshot of the given Agent's current location and heading, the change in these values from the previous step, along with their unique identifier.
\textbf{\textit{``SwarmStates"}} are represented by a collection of states from all member agents along with a variety of metrics such as angular momentum, center of mass, scatter, and radial variance.
Lastly, the \textbf{\textit{``WorldState"}} encompasses the states of all swarms along with all polygons that define the boundaries of the environment.

Besides the three predefined object-state pairs, there are three other notable objects within the system: \textit{\textbf{Sensors}}, \textbf{\textit{Controllers}}, and \textbf{\textit{Visualizers}}.
Each \textbf{\textit{``Sensor"}} is representative of a real-world sensor such as an RGB camera or LIDAR scanner that uses information within the WorldState to recreate a synthetic version of the perspective an Agent would see from their location in the world.
\textbf{\textit{``Controllers"}} accept input from Sensors and modify the location, orientation, and heading of an agent based on their physical dynamics. These dynamics are arbitrary so they can be modified to fit a user's specific application.
\textbf{\textit{``Visualizers"}} in Zespol are separable, optional components of the simulator. They take a WorldState at every time step and generate visual output. We include a visualization system based around Matplotlib \cite{Hunter:2007} to provide users with an example to follow when extending these utilities to support their specific application.

The overall algorithmic flow starts at (1) initializing all Swarms and Agents within the World. (2) The WorldState object is constructed by querying all Swarms and Agents for their SwarmStates and AgentStates, respectively. (3) For every Agent within every Swarm, an artificial sensory perception is calculated in the Agent's Sensor based on its location relative to all other elements in the environment. (4) This perception is then passed to the associated Controller where the AgentState is modified. (5) Once every Agent in every Swarm has calculated their new states, any visualizations and logs can be created. (6) Lastly, the newly accumulated WorldState is used to progress through the next simulation time-step. Figure \ref{fig:program_arch} provides a visual representation for the algorithmic flow between the fundamental Zespol elements. 

\section{Initial Results \& Discussion}
\begin{figure}[h]
    \centering
   \includegraphics[width=0.2\textwidth]{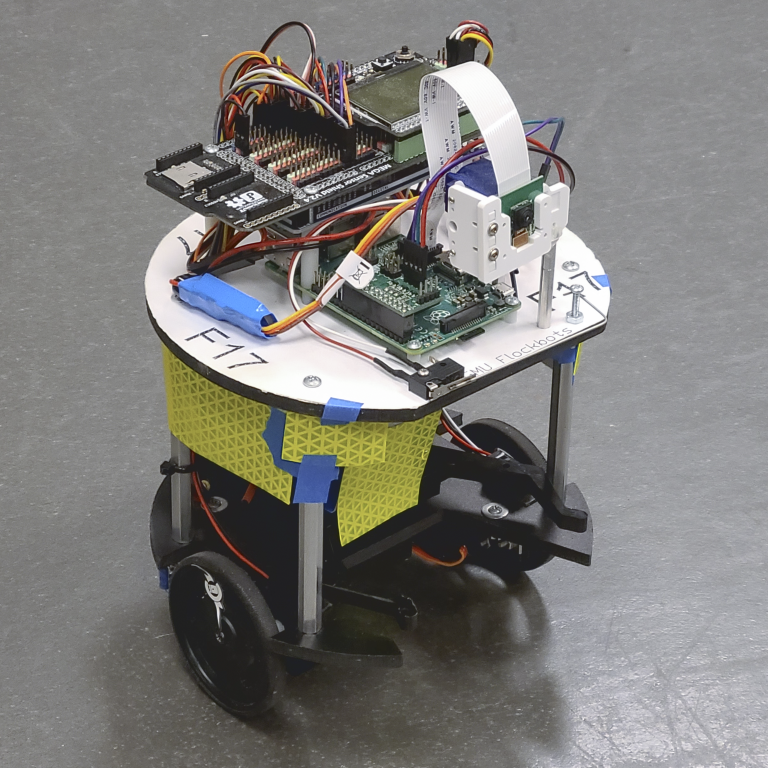}
    \caption{Image of a real-world Flockbot~\cite{SL:KA:MB:DF:KS:BH:CV:AB:RS:BD-14}}
    \label{fig:flockbot}
\end{figure}

Our initial use case for Zespol was recreating the circular milling behavior from \cite{brown2018discovery, https://doi.org/10.48550/arxiv.2301.09018} where agents move in a uniform circle.
Using knowledge gained from \cite{https://doi.org/10.48550/arxiv.2301.09018} and Zespol's modular framework, we set up a simulation environment consisting of 9 Flockbots~\cite{SL:KA:MB:DF:KS:BH:CV:AB:RS:BD-14} with each being equipped with a front-facing infrared proximity sensor and a differential drive system. An image of a real-world Flockbot can be seen in Figure~\ref{fig:flockbot}.

\begin{figure}[h]
    \includegraphics[width=0.4\textwidth]{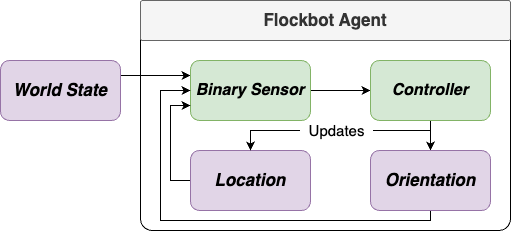}
    \caption{A flowchart presenting a detailed view of the inter-process communication within our example Zespol application using a Flockbot robot with a binary sensor}.
    \label{fig:epuck}
\end{figure}

To fully implement this environment in Zespol, we extended the Agent class with the FlockbotAgent class, a BinarySensor class, and a DifferentialDriveController class. As shown in Figure~\ref{fig:epuck}, the entire process starts at the WorldState going into the BinarySenor where a synthetic binary output is calculated based on the Agent's current location and orientation with respect to the rest of the world. Next, the binary sense is transferred to the DifferentialDriveController where the agent turns left if it senses something or turns right if it senses nothing.

There are numerous parameters for the Flockbot milling behavior that we selected based on the results of \cite{https://doi.org/10.48550/arxiv.2301.09018} where the World ticks at 30 ticks per second, the Swarm contains 9 agents, and Sensors have a view distance of 3 meters and the same asymmetric field-of-view found in \cite{https://doi.org/10.48550/arxiv.2301.09018} with a left bound of 11.5 degrees left of center and a right bound of 4 degrees left of center.

\begin{figure}[h]
    \centering
    \includegraphics[width=0.15\textwidth]{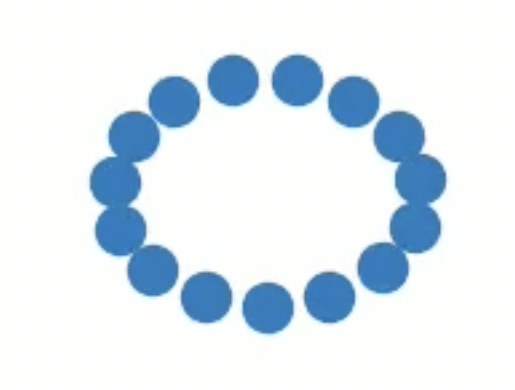}
    {\includegraphics[width=.15\textwidth]{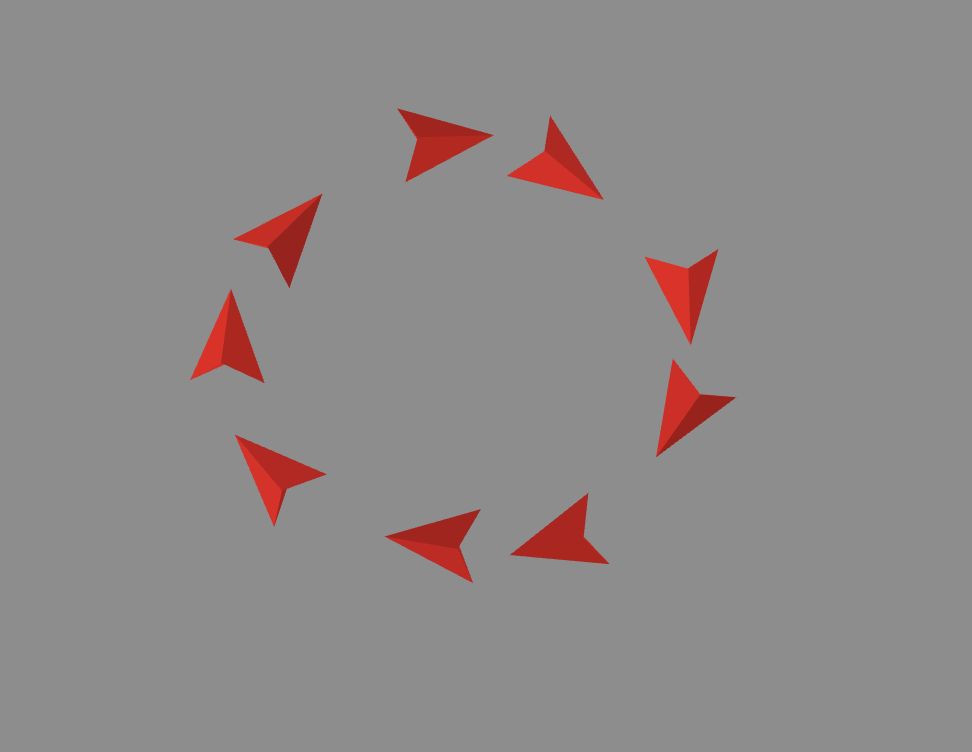}}
    {\includegraphics[width=.15\textwidth]{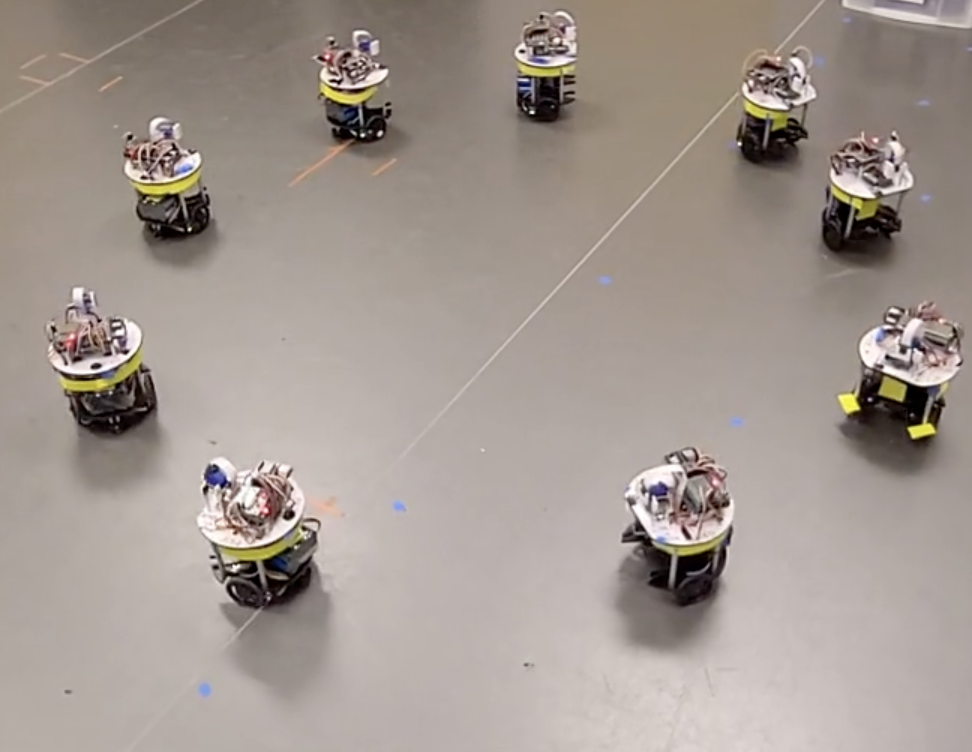}}
    \caption{A swarm of Flockbot robots demonstrating a stable milling behavior in Zespol analogous to results found in NetLogo and real-world experiments}
    \label{fig:epuck-results}
\end{figure}

Zespol successfully models the complex coordination between multiple agents that results in a stable milling behavior. A visualization of the resulting formation is shown in Figure \ref{fig:epuck-results}. This highlights the ability of Zespol to recreate emergent behaviors from other simulated environments and experimental results that have been validated on real-world robotic systems.



\section{Conclusion and Future Work}
In conclusion, the field of agent-based modeling and simulation for studying emergent behaviors has witnessed substantial growth in parallel with the demand for robotic systems that can perform collective tasks.
However, the lack of standardization in simulation environments makes it challenging to compare and contrast newfound research ideas with existing methods.
The Zespol environment is introduced to serve as a lightweight and modular, Python-based simulation environment for developing multi-agent control algorithms. It offers ample opportunities for adoption and expansion by the broader research community. Moreover, the fidelity of Zespol is evaluated against previously published results in simulated and real-world robotics, demonstrating its ability to replicate existing swarming algorithms
with the comparison between Zespol, NetLogo, and real robots conducting the milling behavior with Flockbots. With Zespol, users can develop and standardize swarming algorithms before transitioning over to real-world experiments or higher fidelity simulations. Zespol also provides native support for distributed parallelization across compute clusters and is compatible with neuromorphic computing platforms. As a result, it is a promising solution to issues slowing the advancement of emergent behaviors in robotic swarms of low-powered and individually incapable robotic systems.

Although Zespol is already demonstrating promising results, there is still room for improvement to make it a solid foundation for research on the application of neuromorphic computing in swarming robotics. Our plans include developing formal interfaces for common neuromorphic computing frameworks such as Lava \cite{lava} and Nengo \cite{Bekolay2014}. We will also incorporate formal support for evolutionary algorithms~\cite{coletti2020library} and Bayesian optimization learning schemes~\cite{parsa2021multi}. To simplify the distributed nature of Zespol, we will create a user-friendly interface that minimizes the hassle of dealing with Dask~\cite{dask} and multiprocessing \cite{10.5555/1593511}. Additionally, we will incorporate a vectorized simulation module to run simulations on multiple GPUs across heterogeneous systems. Finally, we will leverage spiking controllers to discover novel swarming behaviors.

\section*{Acknowledgement}
This work was supported in part by the Department of the Navy, Office of Naval Research (ONR), under federal grant N00014-22-1-2207. 

\bibliographystyle{ACM-Reference-Format}
\bibliography{bibliography}

\end{document}